\documentclass[11pt,a4paper]{article}
\usepackage[hyperref]{acl2021}
\usepackage{times}
\usepackage{latexsym}
\usepackage{lineno}
\usepackage{booktabs}
\usepackage{xcolor}

\usepackage{microtype}

\usepackage{enumitem}

\usepackage{linguex} 

\usepackage{tikz}
\usepackage{pgf}
\usetikzlibrary{shapes,external}
\usetikzlibrary{shapes.multipart}
\usetikzlibrary{calc, positioning, automata}

\usepackage{amsmath}
\usepackage{amssymb}
\usepackage{hyperref}
\usepackage{mathtools}
\usepackage{proof}

\usepackage{xcolor}
\definecolor{first}{RGB}{82,82,82}
\definecolor{enc}{RGB}{253,192,134}
\definecolor{dec}{RGB}{127,201,127}
\definecolor{dec2}{RGB}{140,220,140}
\definecolor{emb}{RGB}{190,174,212}
\definecolor{methods}{RGB}{140,22,22}

\usepackage{subcaption}
\usepackage{graphicx}
\usepackage{floatrow}

\usepackage{tabularx}
\usepackage{booktabs}
\usepackage{multicol}
\usepackage{multirow}

\usepackage{url}

\newcolumntype{C}{>{\centering\arraybackslash}X}

\aclfinalcopy 

\title{Fighting the COVID-19 Infodemic with\\ a Holistic BERT Ensemble}

\author{Giorgos Tziafas$^{\diamond}$, Konstantinos Kogkalidis$^{\clubsuit}$, 
 Tommaso Caselli$^{\diamond}$  \\
$^{\diamond}$University of Groningen, $^{\clubsuit}$Utrecht University\\
         Groningen The Netherlands, Utrecht The Netherlands\\
\tt g.tziafas@student.rug.nl,
 k.kogkalidis@uu.nl,
 t.caselli@rug.nl
}

\begin{document}
\maketitle

\begin{abstract}
This paper describes the \textsc{tokofou} system, an ensemble model for misinformation detection tasks based on six different transformer-based pre-trained encoders, implemented in the context of the COVID-19 Infodemic Shared Task for English. We fine tune each model on each of the task's questions and aggregate their prediction scores using a majority voting approach. \textsc{tokofou} obtains an overall F1 score of 89.7\%, ranking first. 

\end{abstract}

\section{Introduction}
Social media platforms, e.g., Twitter, Instagram, Facebook, TikTok among others, are playing a major role in facilitating communication among individuals and sharing of information. 
Social media, and in particular Twitter, are also actively used by governments and health organizations to quickly and effectively communicate key information to the public in case of disasters, political unrest, and outbreaks~\cite{househ2016communicating,stefanidis2017zika,lalone2017embracing,daughton2019identifying,rogers-etal-2019-calls}.

However, there are dark sides to the use of social media. The removal of forms of gate-keeping and the democratization process of the production of information have impacted the quality of the content that becomes available. Misinformation, i.e., the spread of false, inaccurate, misleading information such as rumors, hoaxes, false statements, is a particularly dangerous type of low quality content that affects social media platforms. The dangers of misinformation are best illustrated by considering the combination of three strictly interconnected factors: (i) the diminishing abilities to discriminate between trustworthy sources and information from hoaxes and malevolent agents~\cite{hargittai2010trust}; (ii) a faster, deeper, and broader spread than true information, especially for topics such as disasters and science~\cite{vosoughi2018spread}; (iii) the elicitation of fears and suspicions in the population, threatening the texture of societies. 

The COVID-19 pandemic is the perfect target for misinformation: it is the first pandemic of the Information Age, where social media platforms have a primary role in the information-sphere; it is a natural disaster, where science plays a key role to understand and cure the disease; knowledge about the SARS‑CoV‑2 virus is limited and the scientific understanding is continually developing. To monitor and limit the threats of COVID-19 misinformation, different initiatives have been activated (e.g., \#CoronaVirusFacts Alliance\footnote{\url{https://bit.ly/3uGjwEr}}, EUvsDisinfo\footnote{\url{https://bit.ly/3wPqsBg}}), while social media platforms have been enforcing more stringent policies. Nevertheless, the amount of produced misinformation is such that manual intervention and curation is not feasible, calling for the development of automatic solutions grounded on Natural Language Processing.

The proposed shared task on COVID-19 misinformation presents innovative aspects mirroring the complexity and variation of phenomena that accompanies the spread of misinformation about COVID-19, including fake news, rumors, conspiracy theories, racism, xenophobia and mistrust of science, among others. To embrace the variation of the phenomena, the task organizers have developed a rich annotation scheme based on seven questions~\cite{NLP4IF-2021-COVID19-task}. Participants are asked to design a system capable of automatically labeling a set of messages from Twitter with a binary value (i.e., \textit{yes}/\textit{no}) for each of the seven questions. Train and test data are available in three languages, namely English, Arabic, and Bulgarian. Our team, \textsc{tokofou}, submitted predictions only for the English data by developing an ensemble model based on a combination of different transformer-based pre-trained language encoders. Each pre-trained model has been selected to match the language variety of the data (i.e., tweet) and the phenomena entailed by each of the questions. With an overall F1 score of 89.7 our system ranked first\footnote{Source code is available at \href{https://github.com/gtziafas/nlp4ifchallenge}{https://git.io/JOtpH}.}.

\section{Data}

The English task provides both training and development data. The data have been annotated using a in-house crowdsourcing platform following the annotation scheme presented in~\citet{alam2020fighting}. 

The scheme covers in a very extensive way the complexity of the phenomena that surrounds COVID-19 misinformation by means of seven key questions. The annotation follows a specific pattern after the first question (Q1), that aims at checking whether a message is a verifiable factual claim. In case of a positive answer, the annotator is presented with an additional  set of four questions (Q2--Q5) addressing aspects such as presence of false information, interest for the public, presence of harmful content, and check-worthiness. After this block, the annotator has two further questions. Q6 can be seen as a refinement of the presence of harmful content (i.e, the content is intended to harm society or weaponized to mislead the society), while Q7 asks the annotator whether the message should receive the attention of a government authority. In case of a negative answer to Q1, the annotator jumps directly to Q6 and Q7. Quite interestingly, Q6 lists a number of categories to better identify the nature of the harm (e.g., satire, joke, rumor, conspiracy, xenophobic, racist, prejudices, hate speech, among others). 

The labels of the original annotation scheme present fine-grained categories for each questions, including a \textit{not sure} value. For the task, the set of labels has been simplified to three: \textit{yes}, \textit{no}, and \textit{nan}, with this latter corresponding in some cases to the \textit{not sure} value. Indeed, due to the dependence of Q2--Q5 to a positive answer to Q1, some \textit{nan} values for this set of questions can also correspond to \textit{not applicable} rather than to \textit{not sure}  making the task more challenging than one would expect.

\begin{figure}
    \centering
    \includegraphics[keepaspectratio, width=0.9\columnwidth]{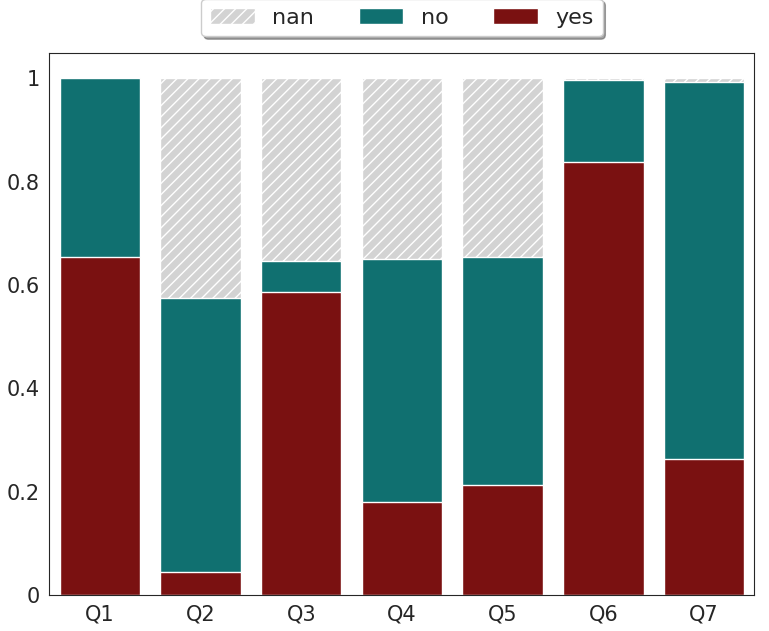}
    \caption{Distribution of the categories for each question in the training data.}
    \label{fig:labels_fig}
\end{figure}

\begin{figure}
    \centering
    \includegraphics[keepaspectratio, width=0.9\columnwidth]{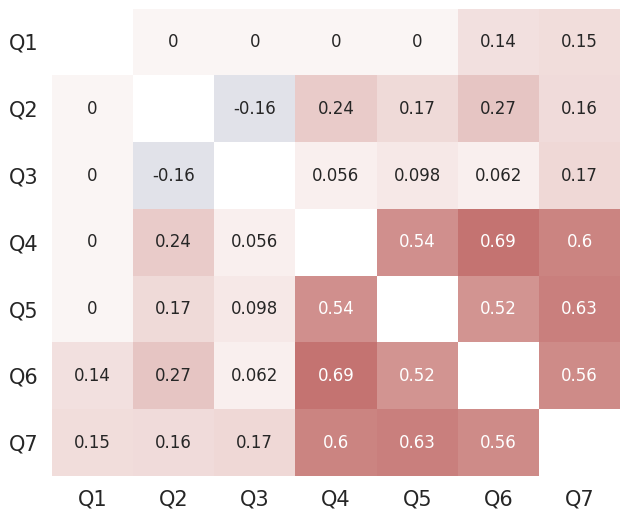}
    \caption{$\phi$ coefficients between question pairs, excluding \textit{nan} values.}
    \label{fig:xcors}
\end{figure}

For English, the organisers released 869 annotated messages for training, 53 for development, and 418 for testing. The distribution of the labels for each question in the training data is reported in Figure~\ref{fig:labels_fig}. As the figures show, the dataset is unbalanced for all questions. While the majority of messages present potential factual claims (Q1), only a tiny minority has been labelled as containing false information (Q2) with a very high portion receiving a \textit{nan} label, suggesting that discriminating whether a claim is false or not is a difficult task for human annotators. Similar observations hold for Q3--Q5. Q6 is a refinement of Q4 about the nature of the harm. The low amount of \textit{nan} values indicates a better reliability of the annotators in deciding the specific type of harms. Q7 also appears to elicit more clear-cut judgements.   
Finally, with the exception of questions Q4--Q7 which exhibit a weak pairwise covariance, no noteworthy correlation is discernible (refer to Figure~\ref{fig:xcors}).

\section{System Overview}
Our system is a majority voting ensemble model based on a combination of six different transformer-based pre-trained encoders, each selected targeting a relevant aspect of the annotated data such as domain, topic, and specific sub-tasks.

\subsection{BERT Models}

Preliminary data analysis and manual inspection of the input texts strongly hint at the notable difficulty of the problem.
The questions our model will be called to answer are high-level semantic tasks that sometimes go beyond sentential understanding, seemingly also relying on external world knowledge.
The limited size of the dataset also rules out the possibility for a task-specific architecture, even more so if one considers the effective loss of data from \textit{nan} labels and the small proportion of development samples, factors that increase the risk of overfitting.
Knowledge grounding with a static external source becomes impractical in view of the rapid pace of events throughout the COVID-19 pandemic: a claim would need to be contrasted against a distinct version of the knowledge base depending on when it was expressed, inserting significant overhead and necessitating an additional timestamp input feature.\footnote{This is especially relevant in the task's context, where the training/development and test data are temporally offset by about a year.}

In light of the above, we turn our attention to pretrained BERT-like models~\cite{bert}.
BERT-like models are the workhorses in NLP, boasting a high capacity for semantic understanding while acting as implicit rudimentary knowledge bases, owing to their utilization of massive amounts of unlabeled data~\cite{petroni-etal-2019-language,bertprimer}.
Among the many candidate models, the ones confined within the twitter domain make for the most natural choices. Language use in twitter messages differs from the norm, in terms of style, length, and content. A twitter-specific model should then already be accustomed to the particularities of the domain, relieving us from either having to account for domain adaptation, or relying on external data.
We obtain our final set of models by filtering our selection in accordance with a refinement of the tasks, as expressed by the questions of the annotation schemes, and the domain. In particular, we focus our selection of models according to the following criteria: (i) models that have been pre-trained on the language domain (i.e, Twitter); (ii) models that have been pre-trained on data related to the COVID-19 pandemic; and (iii) models that have been pre-trained or fine tuned for high-level tasks (e.g., irony and hate speech detection) expressed by any of the target questions. In this way, we identified and used six variations of three 
pre-trained models, detailed in the following paragraphs.

\paragraph{\textsc{BERTweet}}~\cite{vinai} is a RoBERTa\textsubscript{base} model~\cite{roberta} trained from scratch on 850M tweets. 
It is a strong baseline that, fine tuned, achieves state-of-the-art benchmarks on the SemEval 2017 sentiment analysis and the SemEval 2018 irony detection shared tasks~\cite{rosenthal-etal-2017-semeval, van-hee-etal-2018-semeval}.
Here, we use a variant of the model, additionally trained on 23M tweets related to the COVID-19 pandemic, collected prior to September 2020.

\paragraph{\textsc{CT-BERT}}~\cite{del} is a pre-trained BERT\textsubscript{large} model, adapted for use in the twitter setting and specifically the COVID-19 theme by continued unsupervised training on 160M tweets related to the COVID-19 pandemic and collected between January and April 2020.
Fine tuned and evaluated on a small range of tasks, it has been shown to slightly outperform the original.

\paragraph{\textsc{TweetEval}}~\cite{cardiff} is a pre-trained RoBERTa\textsubscript{base} model, further trained with 60M tweets, randomly collected, resulting in a Twitter-domain adapted version. 
We use a selection of four \textsc{TweetEval} models, each fine tuned for a twitter-specific downstream task: hate speech-, emotion- and irony-detection, and offensive language identification.

\subsection{Fine-tuning}
The affinity between the above models and the task at hand allows us to use them for sentence vectorization as-is, requiring only an inexpensive fine tuning pass.
We attach a linear projection on top of each model, which maps its \texttt{[CLS]} token representation to $||Q|| = 7$ outputs, one per question.
The sigmoid-activated outputs act as logits for binary classification and are trained independently with a cross-entropy loss.
We train for $15$ epochs on batches of $16$ tweets, using the \textit{AdamW}~\cite{adamw} optimizer with a learning rate of $3\cdot 10^{-5}$ and weight decay of $0.01$, without penalizing predictions corresponding to \textit{nan} gold labels.
We add dropout layers of rate $0.5$ in each model's classification head.
We perform model selection on the basis of mean F1-score on the development set, and report results in Table~\ref{tab:results}.
As the figures show, no single model outperforms the rest.
Indeed, performance largely varies both across models and questions, with best scores scattered over the table. Similar results occur when repeating the experiments with different random seeds.

\begin{table*}[!h]
    \centering
    \small
    \renewcommand{\arraystretch}{1.05}
    \newcolumntype{M}{>{\centering\arraybackslash}X}
    \newcolumntype{L}{>{\raggedright\arraybackslash}p{0.23\textwidth}}
    \newcolumntype{R}{>{\raggedleft\arraybackslash}p{0.2\textwidth}}
\begin{tabularx}{0.95\textwidth}{@{}LMMMMMMMM@{}}
    {\textbf{Models}} & \textbf{average} & \textbf{Q1} & \textbf{Q2} & \textbf{Q3} & \textbf{Q4} & \textbf{Q5} & \textbf{Q6} & \textbf{Q7} \\
    \toprule
    \textsc{BERTweet } &83.6&86.5&78.4&86.9&88.8&73.4&87.9&\textbf{83.0}\\ 
    \textsc{CT-BERT} &81.3&\textbf{92.4}&76.5&88.5&90.5&68.1&80.5&72.4 \\
    \textsc{TweetEval}-hate &84.8&88.6&84&85.3&90.6&\textbf{82.7}&85.8&70.7 \\
    \textsc{TweetEval}-emotion &84.5&78.2&85.9&\textbf{91.8}&89.0&81.4&85.0&80.0 \\
    \textsc{TweetEval}-irony &\textbf{85.7}&86.5&\textbf{96.1}&85.2&81.6&81.5&\bf88.7&76.7 \\
    \textsc{TweetEval}-offensive &82.9&90.5&74.5&84.1&\textbf{92.2}&72.6&84.4&81.5 \\
    \midrule
    \emph{average} & 83.8 & 87.1 & 82.6 & 87.0 & 88.8 & 76.6 & 85.4 & 77.4\\
    \midrule 
    \emph{Ensemble} &84.6&90.6&78.4&\textbf{91.8}&\textbf{92.2}&76.9&\textbf{90.9}&78.5 \\
    \bottomrule
\end{tabularx}    \caption{Best mean F1-scores (\%) reported in the development set individually for each question as well as their average (with implicit exclusion of \textit{nan} labels for Q2-Q5). Best scores are in bold.}
    \label{tab:results}
\end{table*}

\subsection{Aggregation}
The proposed ensemble model aggregates predictions scores along the 
model axis by first rounding them (into positive or negative labels) and then selecting the final outcome by a majority rule.
The ensemble performs better or equally to all individual models in $3$ out of $7$ questions in the development set, and its metrics lie above the average for $6$ of them.
Keeping in mind the small size of the development set, we refrain from altering the voting scheme, expecting the majority-based model to be the most robust.

During training, we do not apply any pre-processing of the data and rely the respective tokenizer of each model, but homogenize test data by removing URLs.

\section{Results and Discussion}

Results on the test data are illustrated in Table~\ref{tab:score_test}. Two of the three organizers' baselines, namely the majority voting and the ngram baseline, provide already competitive scores. Our ensemble model largely outperforms all of them. The delta with the second best performing system is 0.6 points in F1 score, with a better Recall for \textsc{tokofou} of 3 points.

\begin{table}[!th]
     \centering
     \small
 \begin{tabularx}{0.95\textwidth}{lrrr}
 \textbf{System} & \textbf{Precision} & \textbf{Recall} & \textbf{F1}  \\
  \toprule
 \textsc{tokofou} & 90.7 & 89.6 & 89.7 \\ 
 \emph{Majority Baseline} & 78.6 & 88.3 & 83.0 \\ 
 \emph{Ngram Baseline} & 81.9 & 86.8  & 82.8  \\
 \emph{Random Baseline} & 79.7 & 38.9  & 49.6  \\
 \bottomrule
 \end{tabularx}
 \caption{Results on English test data - average on all questions - and comparison with organizers' baselines.}
     \label{tab:score_test}
 \end{table}

When looking at the results per question,\footnote{Leaderboard is available here: \url{https://tinyurl.com/2drvruc}} \textsc{tokofou} achieves an F1 higher than 90 on Q2 (91.3), Q3 (97.8), and Q6 (90.8). With the exclusion of Q6, the majority baseline on Q2 and Q3 is 92.7 and 100, respectively. This indicates that label imbalance affects the test data as well. At the same time, the performance of ngram baseline suggest that lexical variability is limited. This is not expected given the large variety of misinformation topics that seems affect the discussion around the COVID-19 pandemic. 
These results justify both our choice of models for the ensemble and majority voting as a robust aggregation method.
\section{Conclusion}
We participated in the COVID-19 misinformation shared task with an ensemble of pre-trained BERT-based encoders, fine-tuning each model for predictions in all questions and aggregating them into a final answer through majority voting. 
Our system is indeed a strong baseline for this task showing the effectiveness of available pre-trained language models for Twitter data, mixed with variants fine tuned for a specific topic (COVID-19) and multiple downstream tasks (emotion detection, hate-speech, etc.).  
Results indicate that this holistic approach to transfer learning allows for a data-efficient and compute-conscious methodology, omitting the often prohibitive computational requirement of re-training a model from scratch for a specific task, in favour of an ensemble architecture based on task/domain-similar solutions from a large ecosystem of publicly available models.

With appropriate scaling of the associated dataset, a system as proposed by this paper can be suitably integrated into a human-in-the-loop scenario, serving as an effective assistant in (semi-) automated annotation of Twitter data for misinformation.

\newpage
\section*{Acknowledgments}

The authors would like to acknowledge the RUG university computer cluster, Peregrine, for providing the computational infrastructure which allowed the implementation of the current work.

\bibliographystyle{acl_natbib}
\bibliography{refs}

\end{document}